\documentclass[11pt, a4paper]{article}
\usepackage{amsfonts}
\usepackage{mathrsfs}
\usepackage{amssymb}
\usepackage{amsmath}
\usepackage{graphicx}
\usepackage[flushleft]{caption2}
\makeatletter
  \newcommand\figcaption{\def\@captype{figure}\caption}
\makeatother

\begin{document}

\title{A Study on Effectiveness of Extreme Learning Machine\thanks{The research was supported by
the National Natural Science Foundation of China (No. 60873206), the Natural Science Foundation of Zhejiang Province of
China (No. Y7080235) and the Innovation Foundation of Post-Graduates of Zhejiang Province of China (No.
YK2008066).}}

\author{Yuguang Wang \and Feilong Cao\thanks{Corresponding author: Feilong Cao,  E-mail: \tt feilongcao@gmail.com} \and Yubo Yuan }

 \date{}
\maketitle

\begin{center}
\footnotesize Department of Mathematics,
 China Jiliang University,

 Hangzhou 310018, Zhejiang Province,  P R China

\begin{abstract}
Extreme Learning Machine (ELM), proposed by Huang et al., has been shown a promising learning algorithm for single-hidden layer feedforward neural networks (SLFNs). Nevertheless, because of the random choice of input weights and biases, the ELM algorithm sometimes makes the hidden layer output matrix $\mathbf{H}$ of SLFN not full column rank, which lowers the effectiveness of ELM.  This paper discusses the effectiveness of ELM and proposes an improved algorithm called EELM that makes a proper selection of the input weights and bias before calculating the output weights, which ensures the full column rank of $\mathbf{H}$ in theory. This improves to some extend the learning rate (testing accuracy, prediction accuracy, learning time) and the robustness property of the networks. The experimental results based on both the benchmark function approximation and real-world problems including classification and regression applications show the good performances of EELM.
%{\bf MSC(2000):}  41A17

%{\bf 2000 MR Subject Classification:}\quad  ......
\end{abstract}

\end{center}
\normalsize

\section{Introduction}
Extreme learning machine (ELM) proposed by Huang et al. shows as a useful learning method to train single-hidden layer feedforward neural networks (SLFNs) which have been extensively used in many fields because of its capability of directly approximating nonlinear mappings by input data and providing models for a number of natural and artificial problems that are difficult to cope with by classical parametric techniques. So far there have been many papers addressing relative problems. We refer the reader to \cite{Huang2003}, \cite{Huang_Chen2007}-\cite{Huang_Zhu_Siew2006}, \cite{Mingbin_Huang_Sara_Sun2005}, \cite{Rong_Huang_Sara_Sun2009}, \cite{Zhang_Huang_Sun_Sara2005} and \cite{Zhu_Qin_Suganthan_Huang2005}.

In theory, many researchers have explored the approximation ability of SLFNs.
In 1989, Cybenko \cite{Cybenko1989} and Funahashi \cite{Funahashi1989} proved that any continuous functions can be approximated on a compact set with uniform topology by an SLFN with any continuous, sigmoidal activation function, which made a breakthrough in the artificial neural network field. Leshno \cite{Leshno_Lin_Pinkus_Schocken1993} improved the results of Hornik \cite{Honik1991} and proved that any continuous functions could be approximated by feedforward networks with a nonpolynomial activation function. Furthermore, some deep and systematic studies on  the  condition of activation function can be found in \cite{Chen1994, Chen_Chen1995a, Chen_Chen1995b}.
Recently, Cao et al. \cite{Cao_Xie_Xu2008} constructively gave the  estimation of upper bounds of approximation for continuous functions by SLFNs with the bounded, strictly monotone and odd activation function, which means that the neural networks can be constructed without training as long as the samples are given. In practical applications, for function approximation in a finite training set, Huang and Babri \cite{Huang_Babri1998} showed that an SLFN with at most $N$ hidden nodes and with almost any nonlinear activation function can exactly learn $N$ distinct samples.

In comparison with other traditional learning methods such as BP algorithm, the ELM algorithm proves a faster learning algorithm for SLFNs.  There are some advantages of the ELM algorithm: (1) easy to use and no parameters need to be tuned except predefined network architecture; (2) it is proved a faster learning algorithm compared to other conventional learning algorithms such as BP algorithm. Most training can be accomplished in seconds and minutes (for large-scale complex applications) which might not be easily obtained using other traditional learning methods; (3) it possesses similar high generalization performance as BP and SVM; (4) a wide range of activation functions including all piecewise continuous functions can be used as activation functions in ELM. Among the above four features, the most impressive one is the fast speed of training which is far superior to other conventional learning methods.

However, there are also some shortcomings that the ELM algorithm cannot overcome. Random choosing of input weights and biases easily causes the so called hidden layer output matrix not full column rank. This sometimes makes the linear system that is used to train output weights (linking the hidden layer to the output layer) unsolvable. It also lowers the predicting accuracy. Bartlett \cite{Bartlett1998} pointed out that for feedforward neural networks the smaller the norm of weights and training error are, the better generalization performance the networks tend to have. Therefore, it is necessary to develop a more effective learning method that can overcome this shortcoming and approximate as fast as the ELM algorithm.

This paper tries to design such a learning machine. To achieve the ends, we first properly train the input weights and biases such that the hidden layer output matrix full column rank. Then, a new learning algorithm called effective extreme learning machine (EELM) is proposed, where the strictly diagonally dominant criterion for determining a matrix nonsingular is used to choose the proper input weights and biases. In the first phase of the EELM algorithm the samples are sorted by affine transformation. Due to the assumption of the constructive algorithm, the activation function of the network used in our algorithm should be a Gaussian radial basis-type function. With sorted samples, the Gaussian radial basis-type activation function helps distinguish diagonal elements of the matrix from other non-diagonal ones such that the diagonal elements is larger than the sum of all absolutes of non-diagonal elements. Having chosen input weights and biases properly such that the hidden layer output matrix is full column rank, simple generalized inverse operation gives the output weights.

The difference between the new proposed EELM algorithm and the ELM algorithm lies in the training of input weights and biases. And time spent in the first phase of EELM is very short compared to the second step. So EELM is actually faster than ELM. And EELM algorithm also possesses the qualities of the ELM algorithm including easy implementing, good
generalization performance. Moreover, the new algorithm improved the effectiveness of learning: the full column rank property of the matrix makes the orthogonal projection method, a fast algorithm for solving generalized inverse, available. So, it is called effective extreme learning machine.

This paper is organized as follows. Section~2 gives two theorems that show the two steps in the first phase (training input weights and biases) of EELM algorithm are strictly correct and reasonable theoretically. The constructive proofs in the theorems actually provide the learning method. Section~3 proposes the new EELM learning algorithm for SLFNs. In Section~4, the complexity of the algorithm is given and performance is measured. Section~5 consists of the discussions and conclusions.

\section{Linear inverse problems and Regularization Model of Neural Network}

For $n$ arbitrary distinct samples $\{(X_{i},t_{i})| i=1,2,\dots,n\}$ where $X_{i}=(x_{i1},x_{i2},\dots,x_{id})^{T}\in \mathbb{R}^{d}$ and $t_{i}=(t_{i1},t_{i2},\dots,t_{im})\in \mathbb{R}^{m}$, standard SLFNs with $N$ nodes and activation function $g$ are mathematically modeled as
\begin{equation*}
    G_{N}(X)=\sum_{i=1}^{N}\beta_{i} g(W_{i}\cdot X + b_{i}),
\end{equation*}
here $\beta_{i}=(\beta_{i1},\beta_{i2},\dots,\beta_{iN})^{T}$ is the output weight vector connecting the $i$-th nodes and output nodes, $W_{i}=(w_{i1},w_{i2},\dots,w_{id})^{T}$ is the input weight vector connecting the $i$-th hidden nodes and the input nodes, and the $b_{i}$ is the threshold of the $i$-th hidden node. Approximating the samples with zero error means the proper selection of $\beta_{i}$, $W_{i}$ and $b_{i}$ such that
\begin{equation*}
    \|G_{N}(X_{j})-t_{j}\|=0\quad(j=1,2,\dots,n)
\end{equation*}
or
\begin{equation}\label{chap2eq1}
    G_{N}(X_{j})=t_{j}\quad(j=1,2,\dots,n),
\end{equation}
that is,
\begin{equation*}
    \mathbf{H} \beta = T,
\end{equation*}
here
\begin{eqnarray}
    \mathbf{H}&=&\mathbf{H}(w_{1},w_{2},\dots,w_{N},b_{1},b_{2},\dots,b_{N},X_{1},X_{2},\dots,X_{n})\nonumber\\
    &=& \left(h_{ij}\right)_{n\times n} = \left(\begin{array}{c c@{\;\cdots\;} c}
    g(W_{1}\cdot X_{1}+b_{1})& & g(W_{N}\cdot X_{1}+b_{N})\\
    \vdots  &  & \vdots\\
    g(W_{1}\cdot X_{n}+b_{1}) & & g(W_{N}\cdot X_{n}+b_{N})
    \end{array}\right).\label{chap2eq6}
\end{eqnarray}
As named in Huang et al. \cite{Huang_Babri1998, Huang2003}, $\mathbf{H}$ is called the hidden layer output matrix of the neural networks.

In ELM algorithm, the choice of $\beta_{i}$ and $W_{i}$ is random, which accelerates the rate of learning. Nevertheless, the randomly selection sometimes produces nonsingular hidden layer output matrix which causes no solution of the linear system (\ref{chap2eq1}). To overcome the shortcoming, an extra phase of training the input weights and biases within acceptable steps to keep $\mathbf{H}$ full column rank should be added to the algorithm.

First, we introduce the definition of inverse lexicographical order (or inverse dictionary order) in $\mathbb{R}^{d}$.

\textbf{Definition~2.1.}~~\emph{Suppose $X_{1}, X_{2}\in\mathbb{R}^{d}$ where $X_{i}=(x_{i1},x_{i2},\dots,x_{id})\in \mathbb{R}^{d}$ $(i=1,2)$ are defined as $X_{1}<_{d}X_{2}$ if and only if there exists $j_{0}\in \{1,2,\dots,d\}$ such that $x_{1j_{0}}<x_{2j_{0}}$ and $x_{1j}=x_{2j}$ for $j=j_{0}+1,\dots,d$. $j_{0}$ is called different attribute and denoted by $da(X_{1},X_{2})$ or $da(1,2)$ for convenience if no confusion is produced.}

With the concept of inverse lexicographical order, we obtain the follow theorem that gives a constructive method for sorting high-dimensional vectors via an affine transformation.

\textbf{Theorem~2.2.}~~\emph{For $n$ distinct vectors $X_{1}<_{d}X_{2}<_{d}\dots <_{d}X_{n}\in \mathbb{R}^{d}$ $(d\geq2)$ and $X_{i}=(x_{i1},x_{i2}, \dots, x_{id})^{T}$ such that $\sum_{j=1}^{d}x_{ij}^{2}>0$ for each $i=1,2,\dots,n$. Calculate $W\in \mathbb{R}^{d}$ as follows,
\begin{eqnarray*}
    && w_{j}^{1}=\frac{1}{\max\limits_{i=1,2,\dots,n}\left\{|x_{i j}|\right\}}>0\quad(j=1,2,\dots,d),\\[0.7 em]
    && x_{ij}^{1}=w_{j}^{1}x_{ij}\in [-1,1]\quad(i=1,2,\dots,n,\; j=1,2,\dots,d),\\[0.7 em]
    && y_{ij}^{1}=\left|x_{i+1,j}^{1}-x_{i j}^{1}\right|\quad(i=1,2,\dots,n-1,\; j=1,2,\dots,d),\\[0.7 em]
    && \delta=\log_{10}d+\log_{10}2,\\[0.7 em]
    && n_{j}=\left[-\log_{10}\left(\min_{i=1,2,\dots,n}\left\{y_{ij}^{1}\right\}\right)\right]+\delta\quad(j=1,2,\dots,d),\\[0.7 em]
    && w_{j}^{2}=w_{j}^{1} 10^{\sum_{p=1}^{j}n_{p}}\quad(j=1,2,\dots,d),\\[0.7 em]
    && W=\left(w_{1}^{2},w_{2}^{2},\dots,w_{d}^{2}\right).
\end{eqnarray*}
Then follows
\begin{equation*}
   W\cdot X_{1}<W\cdot X_{2}<\cdots<W\cdot X_{n}.
\end{equation*}
}

\textbf{Proof.}~~For each fixed $i=1,2,\dots,n-1$, set $k_{0}=da(i,i+1)$ which means by Definition~2.1 that $x_{ik_{0}}<x_{i+1,k_{0}}$ and $x_{ij}=x_{i+1,j}$ ($j=k_{0}+1,\dots,d$). Then
\begin{eqnarray*}
  &&W\cdot X_{i+1}-W\cdot X_{i}  \\
  &=& \sum_{k=1}^{k_{0}-1}\left(x_{i+1,k}^{1}-x_{ik}^{1}\right)10^{\sum_{p=1}^{k}n_{p}}+\left(x_{i+1,k_{0}}^{1}-x_{ik_{0}}^{1}\right)10^{\sum_{p=1}^{k_{0}}n_{p}},
\end{eqnarray*}
where by definition $x_{i,k}^{1}, x_{i+1,k}^{1}\in [-1, 1]$ ($k=1,2,\dots,d$) and $x_{i,k_{0}}^{1}< x_{i+1,k_{0}}^{1}$.
Noticing
\begin{eqnarray*}
  10^{\sum_{p=1}^{k}n_{p}} = 10^{\sum_{p=1}^{k_{0}-1}n_{p}}\left(\prod_{p=k+1}^{k_{0}-1}10^{n_{p}}\right)^{-1}
  \leq \frac{10^{\sum_{p=1}^{k_{0}-1}n_{p}}}{d^{(k_{0}-1)-k}}\quad(k=1,2,\dots,k_{0}-1)
\end{eqnarray*}
and
\begin{equation*}
    10^{n_{k_{0}}}\geq10^{-\log_{10}\left(\min\limits_{i=1,2,\dots,n}\left\{y_{ik_{0}}^{1}\right\}\right)+\delta}=\frac{2 d}{\min\limits_{i=1,2,\dots,n}\left\{y_{ik_{0}}^{1}\right\}}.
\end{equation*}
Therefore,
\begin{eqnarray*}
  &&W\cdot X_{i+1}-W\cdot X_{i}  \\
  &\geq& -2\sum_{k=1}^{k_{0}-1}10^{\sum_{p=1}^{k}n_{p}}+\left(y_{ik_{0}}^{1}10^{n_{k_{0}}}\right)10^{\sum_{p=1}^{k_{0}-1}n_{p}}\\
  &\geq& -(2d)\sum_{k=1}^{k_{0}-1}d^{-k} 10^{\sum_{p=1}^{k_{0}-1}n_{p}}+\left(2d\right)10^{\sum_{p=1}^{k_{0}-1}n_{p}}\\
  &=& (2d) 10^{\sum_{p=1}^{k_{0}-1}n_{p}}\left(1-\frac{1-d^{-(k_{0}-1)}}{d-1}\right)>0.
\end{eqnarray*}
This completes the proof of Theorem~2.2.\quad$\Box$

\textbf{Remark~1.}~~\emph{For the case of $d=1$, one needs not to follow the steps of Theorem~2.2 when selecting of $W$. In fact, in the case of $d=1$, the sort operation of the samples $X_{1},X_{2},\dots,X_{n}$ in the inverse lexicographical order via affine transformation can be skipped over. In addition, we don't need the prior sorting of the samples $X_{1},X_{2},\dots,X_{n}$ in the inverse lexicographical order. This is because the computation of $W$ is independent of the order of the samples. Therefore, one only has to sort $W\cdot X_{1}, W\cdot X_{2}, \dots, W\cdot X_{n}$ and change the order of $t_{1},t_{2},\dots,t_{n}$ correspondingly in the linear system.
}

Having calculated the $W$ and given an order to $W\cdot X_{1}, W\cdot X_{2}, \dots, W\cdot X_{n}$, we are able to select input weights and biases, which ensures non-singularity of  the hidden layer output matrix of the neural networks. This is stated in the following theorem. It is noteworthy that the activation function should satisfy some assumptions.

\textbf{Theorem~2.3.}~~\emph{Let $g(x)$ be a positive finite function on $\mathbb{R}$ such that $\lim\limits_{x\rightarrow-\infty}g(x)=\lim\limits_{x\rightarrow+\infty}g(x)=0$ and $g(x)$ is not identically equal to $0$. Given $n$ distinct samples $X_{1}<_{d}X_{2}<\dots <_{d}X_{n}\in \mathbb{R}^{d}$ $(d\geq2)$ and $X_{i}=(x_{i1}, x_{i2}, \dots, x_{id})^{T}$ such that $\sum\limits_{j=1}^{d}x_{ij}^{2}>0$ for any $i=1,2,\dots,n$, there exist input weights $W_{i}\in \mathbb{R}^{d}$ and biases $b_{i}\in \mathbb{R}$ $(i=1,2,\dots,n)$ such that the square matrix
\begin{eqnarray*}
    \mathbf{H}=\left(h_{ij}\right)_{n\times n}=\left(\begin{array}{cc@{\quad\dots\quad}c}
    g(W_{1}\cdot X_{1}+b_{1})& g(W_{2}\cdot X_{1}+b_{2})& g(W_{n}\cdot X_{1}+b_{n})\\
    g(W_{1}\cdot X_{2}+b_{1})& g(W_{2}\cdot X_{2}+b_{2})& g(W_{n}\cdot X_{2}+b_{n})\\
    \vdots & \vdots & \vdots \\
    g(W_{1}\cdot X_{n}+b_{1})& g(W_{2}\cdot X_{n}+b_{2})& g(W_{n}\cdot X_{n}+b_{n})
    \end{array}\right)
\end{eqnarray*}
is nonsingular.}

\textbf{Proof.}~~The proof of the theorem is actually the process of selection of input weights and biases. By assumptions that $g(x)$ is a finite function on $\mathbb{R}$ and $\lim\limits_{x\rightarrow-\infty}g(x)=\lim\limits_{x\rightarrow+\infty}g(x)=0$, $g(x)$ has a maximum on $\mathbb{R}$. Set $M=g(x_{0})=\max\left\{g(x)|x\in\mathbb{R}\right\}$ $(x_{0}\in\mathbb{R})$. For $\varepsilon_{0}=M/n>0$, there exists $a>0$ such that $g(x)<M/n^{2}$ for $|x|>a$ and $x_{0}\in (-a,a)$. Now choose $W=1$ if $d=1$ and $W$ by Theorem~2.2 if $d\geq2$, which implies by assumptions and Theorem~2.2 that
\begin{equation}\label{chap2eq2}
   W\cdot X_{1}<W\cdot X_{2}<\cdots<W\cdot X_{n}.
\end{equation}
Then, select $W_{i}$ and $b_{i}$ ($i=1,2,\dots,n$) as follows.
\begin{eqnarray}
    && dist=\max\{a-x_{0},a+x_{0}\},\label{chap2eq4}\\[0.8 em]
    && k_{i}=\frac{2 dist}{\min\left\{W\cdot X_{i+1}-W\cdot X_{i}, W\cdot X_{i}-W\cdot X_{i-1}\right\}}\quad(i=2,3,\dots,n-1),\label{chap2eq5}\\[0.8 em]
    && W_{i}=k_{i} W\;(i=2,3,\dots,n-1),\quad%\nonumber\\[0.8 em]
    W_{1}=W_{2},\quad%\nonumber\\[0.8 em]
    W_{n}=W_{n-1},\nonumber\\[0.8 em]
    && b_{i}=x_{0}-W_{i}\cdot X_{i}\quad(i=1,2,\dots,n).\label{chap2eq3}
\end{eqnarray}
One thus obtains by (\ref{chap2eq3}) that for $i=1,2,\dots,n$,
\begin{equation*}
    g(W_{i}\cdot X_{i}+b_{i})=M.
\end{equation*}
By (\ref{chap2eq2}), (\ref{chap2eq4}) and (\ref{chap2eq5}), there holds for $i=1,2,\dots,n$,
\begin{equation*}
    \left|\left(W_{i}\cdot X_{j}+b_{i}\right)-\left(W_{i}\cdot X_{i}+b_{i}\right)\right|\geq k_{i}\left|W\cdot X_{j}-W\cdot X_{i}\right|\geq 2\: dist\quad(j=1,2,\dots,n,\;j\neq i),
\end{equation*}
and %by (\ref{chap2eq2}),
\begin{eqnarray*}
  W_{i}\cdot X_{1}+b_{i} &<& W_{i}\cdot X_{2}+b_{i} <\cdots W_{i}\cdot X_{i-1}+b_{i}<x_{0}-a\\
  &<& W_{i}\cdot X_{i}+b_{i}=x_{0}<x_{0}+a<W_{i}\cdot X_{i+1}+b_{i}<\cdots <W_{i}\cdot X_{n}+b_{i}.
\end{eqnarray*}
Therefore,
\begin{equation*}
    g(W_{i}\cdot X_{j}+b_{i})<\frac{M}{n^{2}}.
\end{equation*}
Hence,
\begin{equation*}
    h_{ii}>\sum_{{(k,i)\neq(i,i) \atop k,j=1,2,\dots,n}}|h_{kj}|,
\end{equation*}
thus, $\mathbf{H}$ is strictly diagonally dominant. So $\mathbf{H}$ is nonsingular. This completes the proof of Theorem~2.3.\quad$\Box$

\textbf{Remark~2.}~~\emph{We summarize the steps of selection of input weights and biases as follows. First, choose $a\in\mathbb{R}$ such that
\begin{equation*}
    g(x)<\frac{M}{n^2}\quad(|x|>a,\;x_{0}\in(-a,a)),
\end{equation*}
here,
\begin{equation*}
    M=\max\left\{g(x)|x\in \mathbb{R}\right\}=g(x_{0})\quad(x_{0}\in \mathbb{R}).
\end{equation*}
Then, calculate as follows.
\begin{eqnarray*}
    && dist=\max\{a-x_{0},a+x_{0}\},\\[0.8 em]
    && k_{i}=\frac{2 dist}{\min\left\{W\cdot X_{i+1}-W\cdot X_{i}, W\cdot X_{i}-W\cdot X_{i-1}\right\}}\quad(i=2,3,\dots,n-1),\\[0.8 em]
    && W_{i}=k_{i} W\quad(i=2,3,\dots,n-1),\quad
    W_{1}=W_{2},\quad
    W_{n}=W_{n-1},\\[0.8 em]
    && b_{i}=x_{0}-W_{i}\cdot X_{i}\quad(i=1,2,\dots,n).
\end{eqnarray*}}

\textbf{Remark~3.}~~\emph{Actually, there exist activation functions meeting the conditions of Theorem~2.3. One kind of such activation functions is functions with one peak such as Gaussian radial basis function $g(x)=e^{-x^{2}}$.}

\textbf{Remark~4.}~~\emph{In the case that the number of rows $m$ of $\mathbf{H}$ is larger than that of columns, $W$ and $b$ are calculated based on the square matrix which consists of the $m$ forward rows of $\mathbf{H}$.}

\section{Extreme learning machine using iterative method}
Based upon Theorem~2.2 and Theorem~2.3, a more effective method for training SLFNs is proposed in this section.
\subsection{Features of extreme learning machine (ELM) algorithm}
The ELM algorithm proposed by Huang et al. can be summarized as follows.

\textbf{Algorithm ELM:}~~Given a training set $\mathcal {N}=\{(X_{i},t_{i})|\in \mathbb{R}^{d}, t_{i}\in \mathbb{R}, i=1,2,\dots, n\}$ and activation function $g$, hidden node number $n_{0}$.
\newcounter{marker}
\begin{list}{\emph{Step}~\arabic{marker}:}{\usecounter{marker}
\setlength{\labelwidth}{1.5 cm}\setlength{\leftmargin}{2 cm}
\setlength{\labelsep}{0.3 cm}\setlength{\rightmargin}{1 cm}
\setlength{\parsep}{0.2ex plus0.1ex minus0.1ex}
\setlength{\itemsep}{0ex plus0.1ex}
\setlength{\topsep}{0.1ex plus0.1ex}}
\item Randomly assign input weight $W_{i}$ and bias $b_{i}$\quad($i=1,2,\dots, n_{0}$).
\item Calculate the hidden layer output matrix $\mathbf{H}$.
\item Calculate the output weight $\beta$ by $\beta=\mathbf{H}^{\dag}T$, here $\mathbf{H}^{\dag}$ is the Moore-Penrose generalized inverse of $\mathbf{H}$ (see \cite{Rao_Mitra1972} and \cite{Serre2010} for further details) and $T=(t_{1},t_{2},\dots,t_{n})^{T}$.
\end{list}
The ELM is proved in practice an extremely fast algorithm. This is because it randomly chooses the input weights $W_{i}$ and biases $b_{i}$ of the SLFNs instead of selection. However, this big advantage makes the algorithm less effective sometimes. As mentioned in Section~1, the random choice of input weights and biases is likely to produce an unexpected result, that is, the hidden layer output matrix $\mathbf{H}$ is not full column rank or singular (see (\ref{chap2eq6})), which causes two difficulties. First, it enlarges training error of the samples, which to some extent lowers the prediction accuracy as we can see in the following sections. Besides, the ELM cannot use the orthogonal projection method to calculate Moore-Penrose generalized inverse of $\mathbf{H}$ due to the singularity of $\mathbf{H}$, instead, it prefers singular value decomposition (SVD) which wastes more time.

\subsection{Improvement for the effectiveness of extreme learning machine}

According to Theorem~2.2 and Theorem~2.3, one can summarize the new extreme learning machine for SLFNs as follows. We call the new algorithm effective extreme learning machine. In the algorithm, Gaussian radial basis activation function is used.

\textbf{Algorithm EELM:}~~Given a training data set $\mathcal {N}=\{(X_{i}^{*},t_{i}^{*})|X_{i}^{*}\in \mathbb{R}^{d}, t_{i}^{*}\in \mathbb{R}, i=1,2,\dots, n\}$, activation function of radial basis function $g(x)=e^{-x^{2}}$ and hidden node number $n_{0}$.

\emph{Step}~1:~~Select weights $W_{i}$ and bias $b_{i}$ $(i=1,2,\dots,n_{0})$.

Sort $W\cdot X_{1}^{*}, W\cdot X_{2}^{*}, \dots, W\cdot X_{n_{0}}^{*}$ in order that $W\cdot X_{i_1}^{*}<W\cdot X_{i_2}^{*}<\dots< W\cdot X_{i_{n_{0}}}^{*}$ ($i_{j}\neq i_{k}$ for $j\neq k$, $j,k=1,2,\dots,n_{0}$ and $i_{j}=1,2,\dots,n_{0}$) are satisfied, then correspondingly change the order of the forward $n_{0}$ samples $(X_{i}^{*},t_{i}^{*})$ ($i=1,2, \dots, n_{0}$). And denote the sorted data by $(X_{i},t_{i})$ ($i=1,2,\dots,n$) and $X_{i}=(x_{i 1}, x_{i 2}, \dots, x_{i d})$ $(i=1,2,\dots,n)$. For $j=1,2,\dots,d$, make following calculations.
\begin{eqnarray*}
    && w_{j}^{1}=\frac{1}{\max\limits_{i=1,2,\dots,n_{0}}\left\{|x_{i j}|\right\}}>0,\\[0.8 em]
    && x_{ij}^{1}=w_{j}^{1}x_{ij}\in [-1,1],\\[0.8 em]
    && y_{ij}^{1}=\left|x_{i+1,j}^{1}-x_{i j}^{1}\right| \quad(i=1,2,\dots,n_{0}-1),\\[0.8 em]
    && \delta=\log_{10}d+\log_{10}2,\\[0.7 em]
    && n_{j}=\left[-\log_{10}\left(\min_{i=1,2,\dots,n}\left\{y_{ij}^{1}\right\}\right)\right]+\delta,\\[0.8 em]
    && w_{j}^{2}=w_{j}^{1} 10^{\sum_{p=1}^{j}n_{p}}.
\end{eqnarray*}
Set
\begin{equation*}
    W=\left(w_{1}^{2},w_{2}^{2},\dots,w_{d}^{2}\right).
\end{equation*}
Let
\begin{eqnarray*}
    M=1,\quad
    x_0=0,\quad
    a=\max\left\{\sqrt{\left|2\ln(n_{0})\right|},1\right\}+1
\end{eqnarray*}
such that
$M=\max\left\{g(x)|x\in \mathbb{R}\right\}=g(x_{0})\;(x_{0}\in \mathbb{R})$ and $g(x)<M/n_{0}^2\;(|x|>a,\;x_{0}=0\in(-a,a))$.
Then,
\begin{eqnarray*}
    && dist=\max\{a-x_{0},a+x_{0}\},\\[0.8 em]
    && k_{i}=\frac{2 dist}{\min\left\{W\cdot X_{i+1}-W\cdot X_{i}, W\cdot X_{i}-W\cdot X_{i-1}\right\}}\quad(i=2,3,\dots,n_{0}-1),\\[0.8 em]
    && W_{i}=k_{i} W\quad(i=2,3,\dots,n_{0}-1),\quad%\\[0.8 em]
    W_{1}=W_{2},\quad%\\[0.8 em]
    W_{n_{0}}=W_{n_{0}-1},\\[0.8 em]
    && b_{i}=x_{0}-W_{i}\cdot X_{i}\quad(i=1,2,\dots,n_{0}).
\end{eqnarray*}

\emph{Step}~2:~~Calculate output weights $\beta=(\beta_{1},\beta_{2},\dots,\beta_{n_{0}})$ $(i=1,2,\dots,n_{0})$.

Let $T=(t_{1},t_{2},\dots,t_{n})^{T}$ and
\begin{eqnarray*}
    \mathbf{H}=\left(\begin{array}{cc@{\quad\dots\quad}c}
    g(W_{1}\cdot X_{1}+b_{1})& g(W_{2}\cdot X_{1}+b_{2})& g(W_{n_{0}}\cdot X_{1}+b_{n_{0}})\\
    %g(W_{1}\cdot X_{2}+b_{1})& g(W_{2}\cdot X_{2}+b_{2})& g(W_{n}\cdot X_{2}+b_{n})\\
    \vdots & \vdots & \vdots \\
    g(W_{1}\cdot X_{n_{0}}+b_{1})& g(W_{2}\cdot X_{n_{0}}+b_{2})& g(W_{n_{0}}\cdot X_{n_{0}}+b_{n_{0}})\\
    g(W_{1}\cdot X_{n_{0}+1}+b_{1})& g(W_{2}\cdot X_{n_{0}+1}+b_{2})& g(W_{n_{0}}\cdot X_{n_{0}+1}+b_{n_{0}})\\
    \vdots & \vdots & \vdots \\
    g(W_{1}\cdot X_{n}+b_{1})& g(W_{2}\cdot X_{n}+b_{2})& g(W_{n_{0}}\cdot X_{n}+b_{n_{0}})\\
    \end{array}\right)_{n\times n_{0}}.
\end{eqnarray*}
Then
\begin{equation*}
    \beta=\mathbf{H}^{\dag} T=(\mathbf{H}^{T}\mathbf{H})^{-1}\mathbf{H}^{T}T.
\end{equation*}

\textbf{Remark~5.}~~\emph{As pointed out in Remark~3, the activation function in the EELM algorithm for SLFNs should be chosen to satisfy the assumption of Theorem~2.3. By Theorem~2.3, if the samples possess the properties that they are distinct and $\sum\limits_{j=1}^{d}x_{ij}^{2}>0$ (this is actually almost surely), then the hidden layer output matrix $\mathbf{H}$ is full column rank. This ensures that $\mathbf{H}^{T}\mathbf{H}$ is nonsingular and thus the fast orthogonal project method can be used in computation of Moore-Penrose generalized inverse. Moreover, when $n=n_{0}$, that is, $\mathbf{H}$ is an invertible square matrix.}

\textbf{Remark~6.}~~\emph{In accordance with Theorem~2.3, $X_{1}<_{d}X_{2}<_{d}\cdots <_{d}X_{n_{0}}$, here the $(X_{i},t_{i})$ $(i=1,2,\dots,n)$ denotes the sorted samples as in the algorithm above. To avoid the random error, the $n_{0}$ samples being used to train input weights and biases can be randomly chosen from the original $n$ samples.}

\section{Complexity and performance}
The proposed EELM spends more time on training samples than ELM but in fact the extra time spent on selecting input weights and biases is $\mathcal{O}(n_{0}d)$. Compared with the second phase of the algorithm that calculates the output weights, it can be viewed as a constant.

In the rest part of this section, the performance of the proposed EELM learning algorithm is measured compared with the ELM algorithm. The simulations for ELM and EELM algorithms are carried out in the Matlab 7.0 environment running in Intel Celeron 743 CPU with the speed of $1.30$ GHz and in Intel Core 2 Duo CPU. The activation function used in both algorithm is Gaussian radial basis function $g(x)=e^{-x^{2}}$.

\subsection{Benchmarking with a regression problem: approximation of `SinC' function with noise}
First of all, we use the `Sinc' function to measure the performance of the two algorithms. The target function is as follows.
\begin{eqnarray*}
    y=f(x)=\left\{\begin{array}{ll}
    \sin(x)/x & x\neq0,\\
    1 & x=0.
    \end{array}\right.
\end{eqnarray*}
A training set $(X_i, t_i)$ and testing set $(X_i, t_i)$ with $200$ samples are respectively created, where $X_i$ in the training data is distributed in $[-10,10]$ with uniform step length. The $X_i$ in the testing data is chosen randomly in the standard normalized distribution in $[-30,30]$. The reason why the range $([-30,30])$ of testing data is longer than that of training data is because an obvious way to assess the quality of the learned model is to see on how long term the predictions given by the model are accurate.
The experiment is carried out on these data as follows. There are $200$ hidden nodes assigned for both the ELM and the EELM algorithms. $50$ trials have been conducted for the ELM algorithm to eliminate the random error and the results shown are the average results. Results shown in Table~\ref{chap4tab1} include training time, testing time, training accuracy, testing accuracy and the number of nodes of both algorithms.

\begin{table}
\centering
\setlength{\abovecaptionskip}{2pt}%
\setlength{\belowcaptionskip}{8pt}%
\renewcommand{\captionstyle}{flushleft}
\renewcommand{\captionlabeldelim}{.~}
\renewcommand{\captionfont}{\footnotesize \sffamily}
\renewcommand{\captionlabelfont}{\footnotesize \sffamily}
\caption{Performance comparison for learning function: SinC}\label{chap4tab1}
\mbox{\footnotesize
\begin{tabular}{@{}c@{\;\;\quad}llll@{\;\;\quad}lll@{\;\;\quad}l@{}}
  \hline
  Algorithms&  & Time    &         & & Accuracy              &           & &  No. of Nodes \\ \cline{3-4} \cline{6-7}
            &  &Training & Testing & &   Training            &  Testing  & &    \\ \hline
  ELM       &  &$0.0870$ & $0.0056$  & & $3.1431\times10^{-6}$ & $5.6642$    & &  $200$ \\
  EELM      &  &$0.0624$ &  $0 $     & & $0.0038$                    & $0.1595$    & &  $200$ \\
  % &  &  &  &  &  &  &  \\
  \hline
\end{tabular}}
\end{table}

\begin{center}
\begin{minipage}[t]{0.48\textwidth}
  \renewcommand{\captionlabeldelim}{.~}
  \renewcommand{\captionfont}{\scriptsize \sffamily}
  \renewcommand{\captionlabelfont}{\scriptsize \sffamily}
  \vspace{0pt}
  \centering
  \includegraphics[scale=0.46]{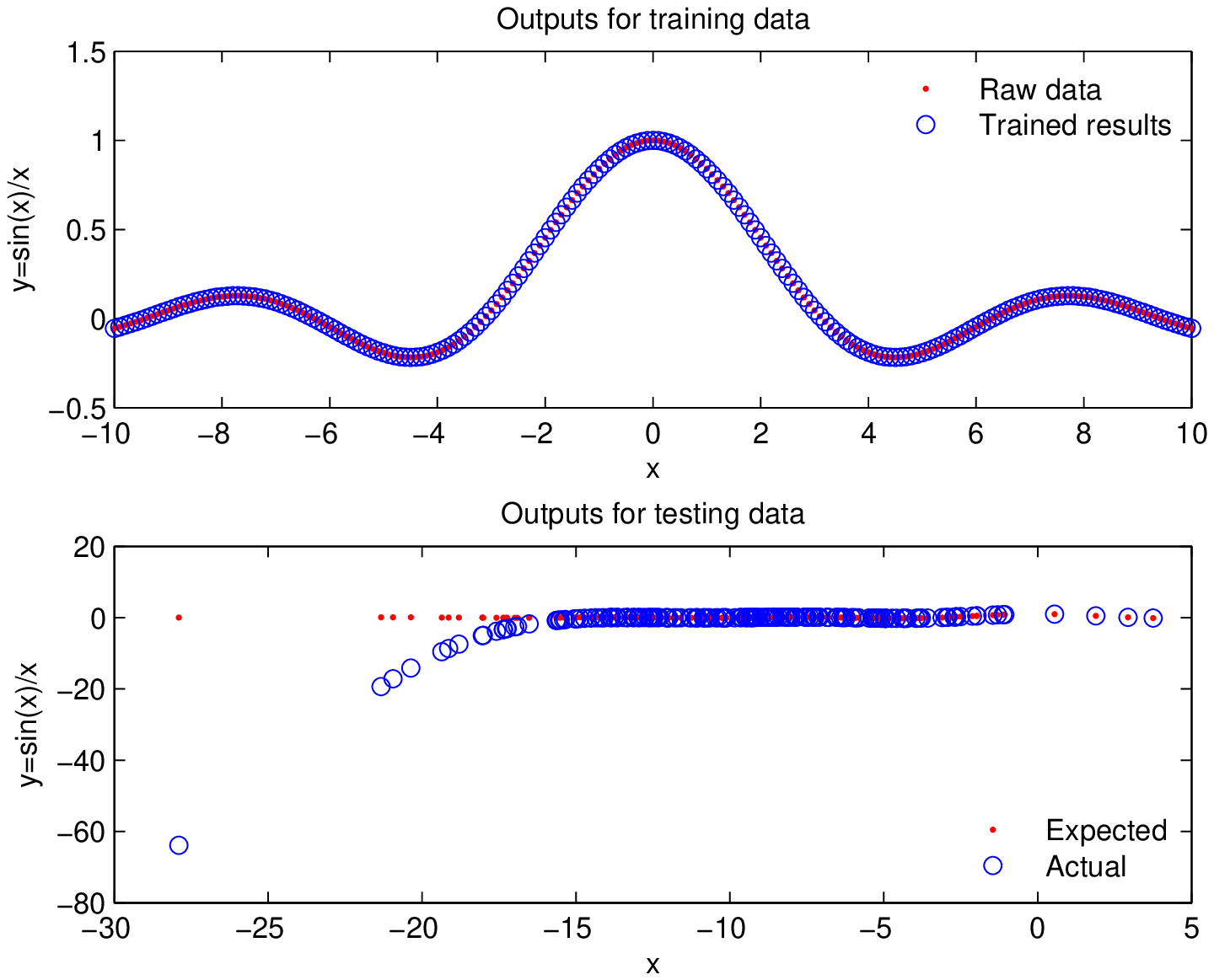}
  %\vspace{3ex}
  \renewcommand{\figurename}{Fig.}\vspace{-7ex}
  \figcaption{Outputs of ELM learning algorithm}\label{fig:SincELM}
\end{minipage}
\begin{minipage}[t]{0.48\textwidth}
   %Requires \usepackage{graphicx}
  \renewcommand{\captionlabeldelim}{.~}
  \renewcommand{\captionfont}{\scriptsize \sffamily}
  \renewcommand{\captionlabelfont}{\scriptsize \sffamily}%\vspace{-1ex}
  \vspace{0pt}
  \centering
  \includegraphics[scale=0.46]{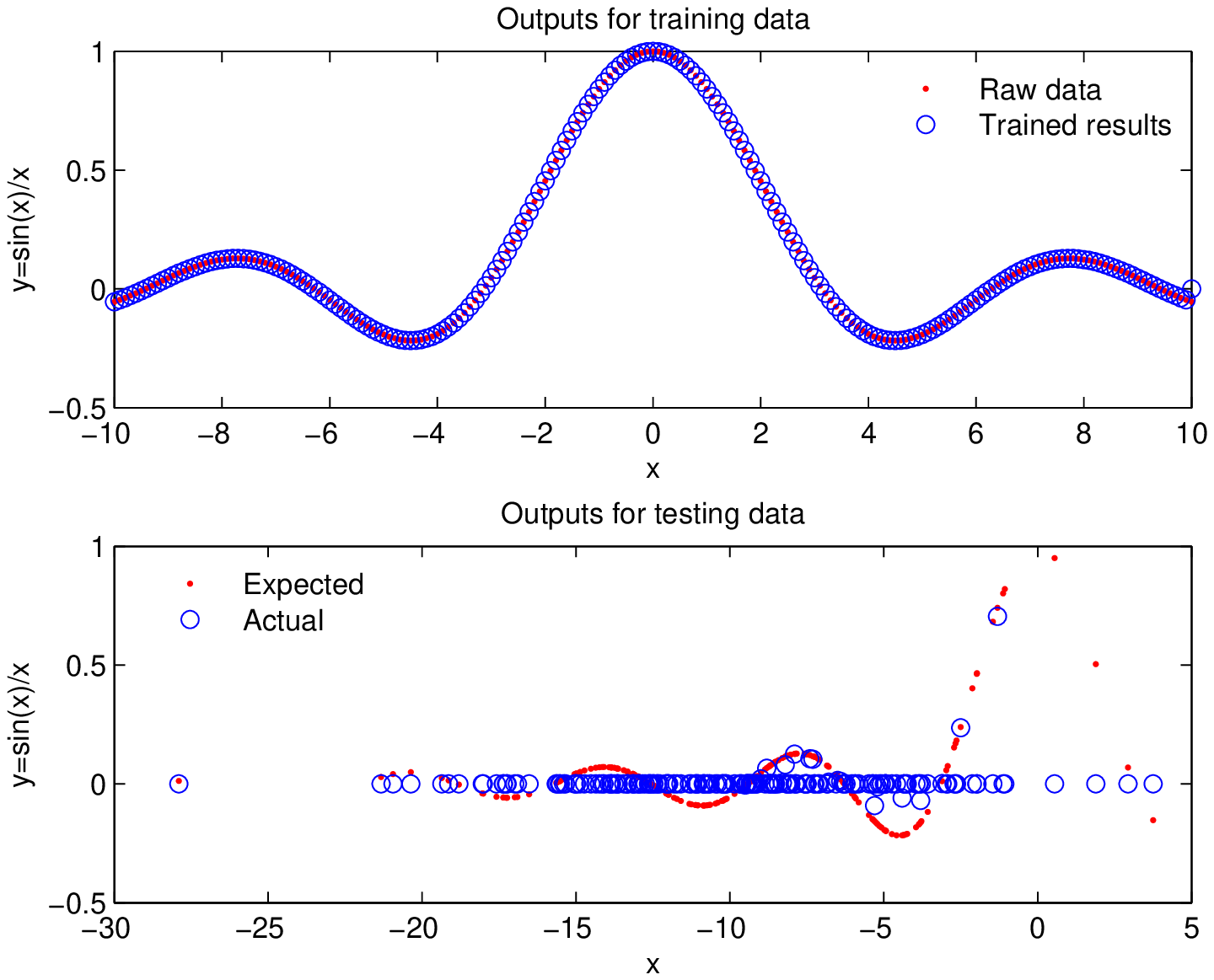}
  %\vspace{0pt}
  \renewcommand{\figurename}{Fig.}\vspace{-7ex}
  \figcaption{Outputs of EELM learning algorithm}\label{fig:SincEELM}
\end{minipage}
\end{center}
It can be seen from Table~\ref{chap4tab1} that the EELM algorithm spent $0.0624s$ CPU time obtaining testing accuracy $0.1595$ with zero training error whereas it takes  $0.0870s$ CPU time for the ELM algorithm to reach a higher error $5.6642$ with training error of $3.1431\times10^{-6}$. Fig.~\ref{fig:SincEELM} shows the expected and approximated results of EELM algorithm and Fig.~\ref{fig:SincELM} shows the true and the approximated results of ELM algorithm. The results show that our EELM algorithm can not only approximate the training data with zero error but also have a long term prediction accuracy. And the time consumption is not more than the ELM algorithm. Though the ELM algorithm has good performance in the interval $[-10,10]$, its long term prediction accuracy is not very satisfactory.

\subsection{Benchmarking with real-world applications}
In this section, we conduct the performance comparison of the proposed EELM and the ELM algorithms for 5 real problems: 3 classification tasks including Diabetes, Glass Identification (Glass ID), Statlog (Landsat Satellite), and 2 regression tasks including Housing and Slump (Concrete Slump). All the data sets are from UCI repository of machine learning databases \cite{Blake_Merz1998}. The speculation of each database is shown in the Table~\ref{tab:EELMSpeculation}. For the databases that have only one data table, as conducted in \cite{Freund_Schapire1996, Ratsch_Onoda_Muller1998, Romero2002, Wilson1996}, $75\%$ and $25\%$ of samples in the problem are randomly chosen for training and testing respectively at each trial.

Fifty trials were conducted for the two algorithms and the results are reported in Table~\ref{tab:EELMAccuracy}, Table~\ref{tab:EELMSTD} and Table~\ref{tab:EELMTime}, which show that in our simulation, generally speaking, ELM can reach higher testing rate for mid and large size classification problems than EELM and for the small size ones, EELM can achieve a higher rate than ELM. In the regression cases, EELM has a better robustness property than ELM. Fig.~\ref{fig:HousingTrain}, Fig.~\ref{fig:HousingTest} and Fig.~\ref{fig:HousingTime} show that EELM is more steady than ELM for regression cases.
\vspace{3ex}
% EELMSpeculation
\begin{table}[htbp]
  \centering
  \setlength{\abovecaptionskip}{2pt}%
\setlength{\belowcaptionskip}{8pt}%
\renewcommand{\captionstyle}{flushleft}
\renewcommand{\captionlabeldelim}{.~}
\renewcommand{\captionfont}{\footnotesize \sffamily}
\renewcommand{\captionlabelfont}{\footnotesize \sffamily}
  \caption{Speculations of real-world applications and the number of nodes for each}%\vspace{1ex}
    \mbox{\footnotesize\begin{tabular}{l@{\quad}llllll}
    %\addlinespace
    %\toprule
    \hline
    \multicolumn{ 1}{c}{Data sets} & \multicolumn{ 2}{c}{\# Observations} &       & \# Attributes & Associated Tasks & \#Nodes \\ \cline{2-3}
    %\midrule
    \multicolumn{ 1}{c}{} & Training & Testing &       & Continuous &       &  \\ \hline
    Diabetes & $576$  & $192$  &       & $8$    & Classification & $20$ \\
    Statlog & $4435$  & $2000$  &       & $36$    & Classification & $20$ \\
    Glass ID & $160$   & $54$    &       & $9$     & Classification & $10$ \\
    Housing & $378$   & $126$   &       & $14$    & Regression & $80$ \\
    Slump & $76$    & $27$    &       & $10$    & Regression & $10$ \\ \hline
    %\bottomrule
    \end{tabular}}
  \label{tab:EELMSpeculation}
\end{table}
\vspace{3ex}
% EELM_Accuracy
\begin{table}[htbp]
  \centering
  \setlength{\abovecaptionskip}{2pt}%
\setlength{\belowcaptionskip}{8pt}%
\renewcommand{\captionstyle}{flushleft}
\renewcommand{\captionlabeldelim}{.~}
\renewcommand{\captionfont}{\footnotesize \sffamily}
\renewcommand{\captionlabelfont}{\footnotesize \sffamily}
  \caption{Comparison training and testing accuracy (error) of ELM and EELM}%\vspace{1ex}
    \mbox{\footnotesize\begin{tabular}{l@{\quad \quad \quad}l@{\quad \quad}l@{\quad \quad \quad}l@{\quad}l@{\quad \quad \quad}l}
    %\addlinespace
    %\toprule
    \hline
    \multicolumn{ 1}{c}{Data sets} & \multicolumn{ 2}{c}{ELM} &     & \multicolumn{ 2}{c}{EELM} \\ \cline{2-3} \cline{5-6}
    %\midrule
    \multicolumn{ 1}{c}{} & Training & Testing &     & Training & Testing \\ \hline
    Diabetes & $0.7703$  & $0.7772$  &     & $0.6767$  & $0.6881$  \\
    Statlog & $0.8159$  & $0.8125$  &     & $0.4711$  & $0.4652$  \\
    Glass ID & $0.9483$  & $0.4219$  &     & $0.9109$  & $0.4489$  \\ %\hline
    Housing & $7.4639$ & $1.1198\times 10^{10}$  &     & $22.7959$  & $15.3878$  \\
             &         &   326.5156 (best)                       &     &            &    \\
    Slump & $7.9927$  & $3.2439\times10^{5}$ &     & $17.8965$  & $19.1654$  \\
             &         &   8.8406 (best)   &     &            &    \\ \hline
    %\bottomrule
    \end{tabular}}
  \label{tab:EELMAccuracy}
\end{table}
\vspace{3ex}

% EELM_STD
\begin{table}[htbp]
  \centering
  \setlength{\abovecaptionskip}{2pt}%
\setlength{\belowcaptionskip}{8pt}%
\renewcommand{\captionstyle}{flushleft}
\renewcommand{\captionlabeldelim}{.~}
\renewcommand{\captionfont}{\footnotesize \sffamily}
\renewcommand{\captionlabelfont}{\footnotesize \sffamily}
  \caption{Comparison of training and testing RMSE of ELM and EELM}%\vspace{1ex}
    \mbox{\footnotesize\begin{tabular}{l@{\quad \quad \quad}l@{\quad \quad \quad}l@{\quad \quad \quad}l@{\quad}l@{\quad \quad \quad}l}
    %\addlinespace
    %\toprule
    \hline
    \multicolumn{ 1}{l}{Data sets} & \multicolumn{ 2}{c}{ELM} &       & \multicolumn{ 2}{c}{EELM} \\ \cline{2-3} \cline{5-6}
    %\midrule
    \multicolumn{ 1}{c}{} & Training & Testing &       & Training & Testing \\ \hline
    Diabetes & $0.0104$  & $0.0288$  &     & $0.0252$  & $0.0374$  \\
    Statlog & $0.0114$  & $0.0104$  &       & $0.0508$  & $0.0525$  \\
    Glass ID & $0.0028$  & $0.0286$  &       & $0.0339$  & $0.0258$  \\
    Housing & $0.2381$  & $3.6669\times 10^{10}$  &     & $0.5161$  & $0.7257$  \\
    Slump & $0.1394$  & $2.1991\times 10^{6}$ &      & $0.5151$  & $0.6818$  \\
    \hline
    %\bottomrule
    \end{tabular}}
  \label{tab:EELMSTD}
\end{table}
\vspace{3ex}

% EELM_Time
\begin{table}[htbp]
  \centering
  \setlength{\abovecaptionskip}{2pt}%
\setlength{\belowcaptionskip}{8pt}%
\renewcommand{\captionstyle}{flushleft}
\renewcommand{\captionlabeldelim}{.~}
\renewcommand{\captionfont}{\footnotesize \sffamily}
\renewcommand{\captionlabelfont}{\footnotesize \sffamily}
  \caption{Comparison of average training and testing time of ELM and EELM}%\vspace{1ex}
    \mbox{\footnotesize\begin{tabular}{l@{\quad \quad \quad}l@{\quad \quad \quad}l@{\quad \quad \quad}l@{\quad \quad}l@{\quad \quad \quad}l}
    %\addlinespace
    %\toprule
    \hline
    \multicolumn{ 1}{l}{Data sets} & \multicolumn{ 2}{c}{ELM(s)} &       & \multicolumn{ 2}{c}{EELM(s)} \\ \cline{2-3} \cline{5-6}
    %\midrule
    \multicolumn{ 1}{c}{} & Training & Testing &       & Training & Testing \\ \hline
    Diabetes & $0.0034$  & $0.0012$  &     & $0.0094$  & $0.0012$  \\
    Statlog & $0.0208$  & $0.0208$  &       & $0.0271$  & $0.0083$  \\
    Glass ID & $0.0012$  & $0.0006$  &       & $0.0246$  & $0.0012$  \\
    Housing & $0.0508$  & $0.0117$  &       & $0.0258$  & $0.0023$  \\
    Slump & $6.2500\times10^{-4}$  & $0.0022$  &       & $0.0075$  & $0.0028$  \\ \hline
    %\bottomrule
    \end{tabular}}
  \label{tab:EELMTime}
\end{table}
\vspace{3ex}
%% Housing Database
\begin{center}
\begin{minipage}[t]{\textwidth}
  \renewcommand{\captionlabeldelim}{.~}
  \renewcommand{\captionfont}{\footnotesize \sffamily}
  \renewcommand{\captionlabelfont}{\footnotesize \sffamily}
  \vspace{0pt}
  \centering
  \includegraphics[scale=0.8]{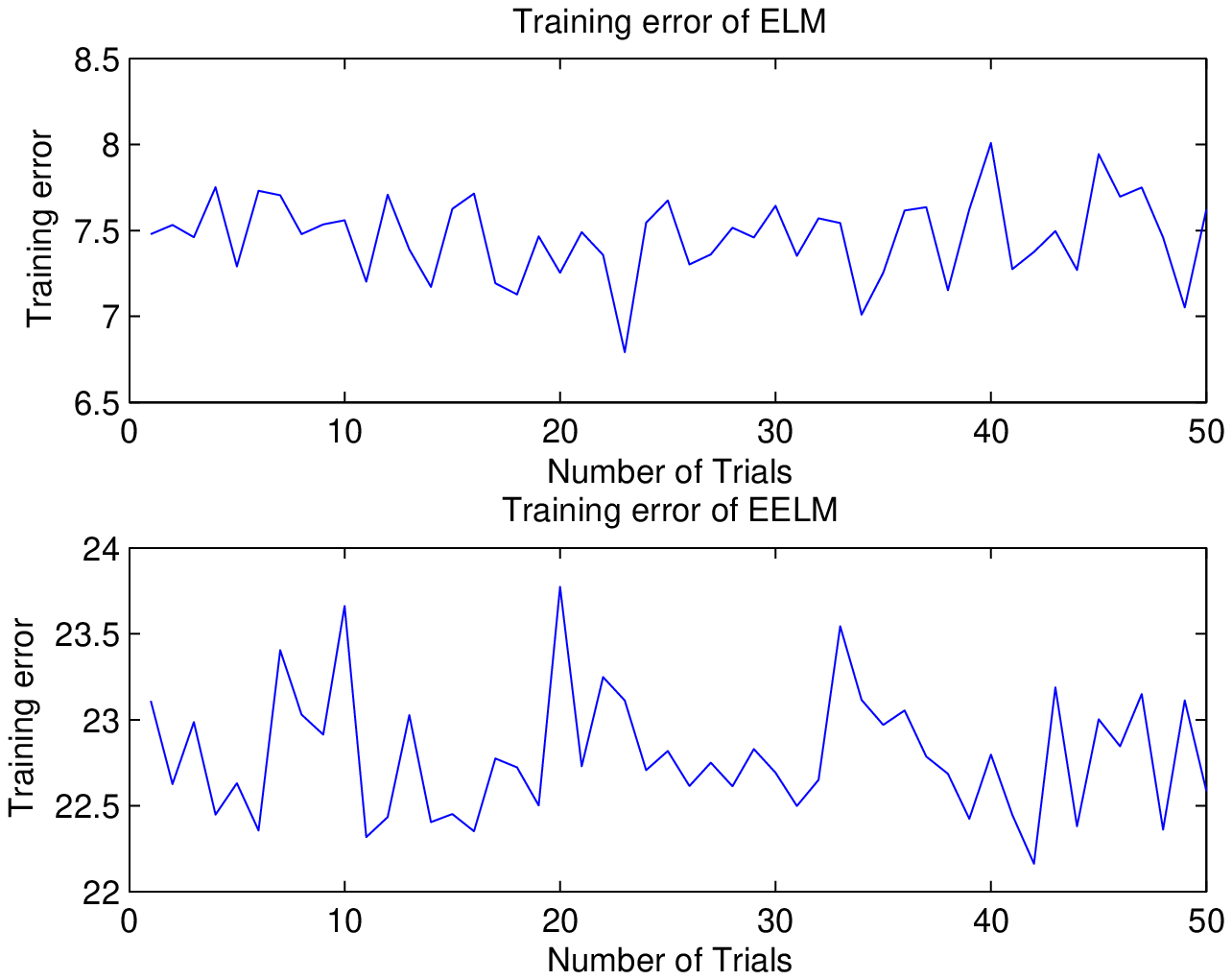}
  \vspace{0ex}
  \renewcommand{\figurename}{Fig.}\vspace{-3ex}
  \figcaption{Training accuracy of two algorithms for Housing}\label{fig:HousingTrain}
\end{minipage}%\hfill
\vspace{5ex}
\begin{minipage}[t]{\textwidth}
   %Requires \usepackage{graphicx}
  \renewcommand{\captionlabeldelim}{.~}
  \renewcommand{\captionfont}{\footnotesize \sffamily}
  \renewcommand{\captionlabelfont}{\footnotesize \sffamily}%\vspace{-1ex}
  \vspace{0pt}
  \centering
  \includegraphics[scale=0.8]{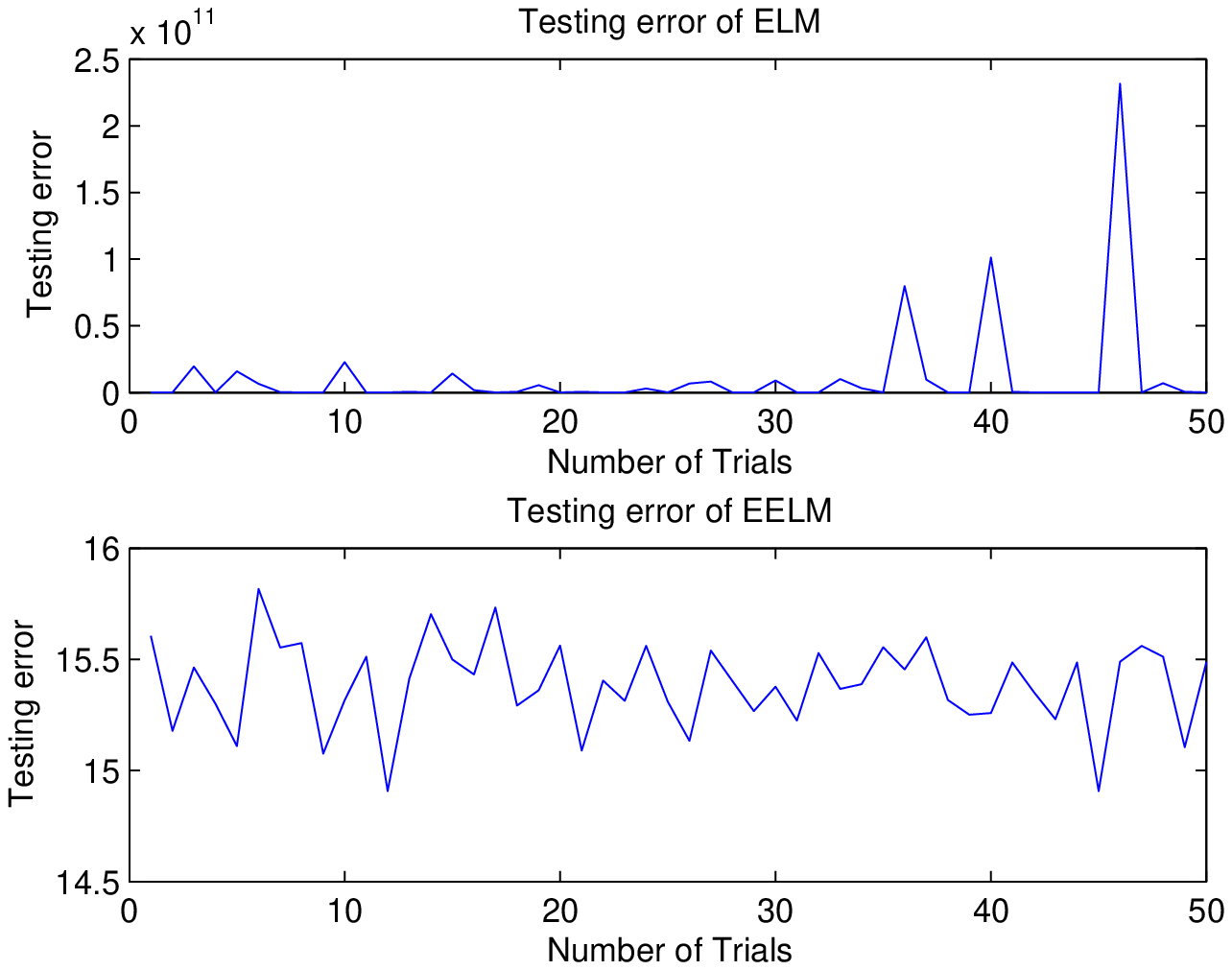}
  \vspace{-0ex}
  \renewcommand{\figurename}{Fig.}\vspace{-3ex}
  \figcaption{Testing accuracy of two algorithms for Housing}\label{fig:HousingTest}
\end{minipage}%\hfill
\vspace{5ex}
\begin{minipage}[t]{\textwidth}
   %Requires \usepackage{graphicx}
  \renewcommand{\captionlabeldelim}{.~}
  \renewcommand{\captionfont}{\footnotesize \sffamily}
  \renewcommand{\captionlabelfont}{\footnotesize \sffamily}%\vspace{-1ex}
  \vspace{0pt}
  \centering
  \includegraphics[scale=0.8]{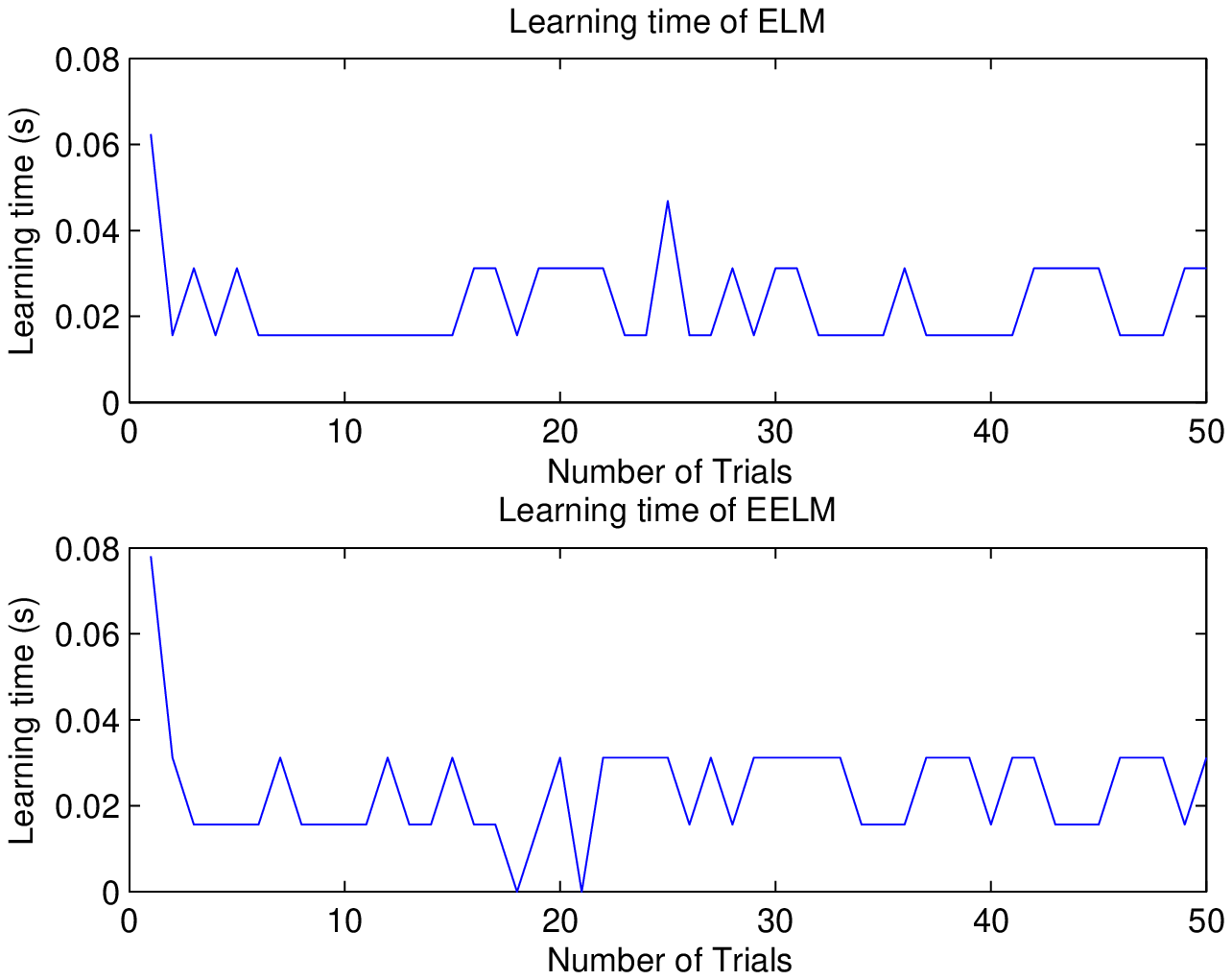}
  \vspace{-0ex}
  \renewcommand{\figurename}{Fig.}\vspace{-3ex}
  \figcaption{Learning time of two algorithms for Housing}\label{fig:HousingTime}
\end{minipage}\hfill
\vspace{1ex}
\end{center}

In the Diabetes case, the performance of both ELM and EELM including training accuracy, testing accuracy and learning time of two algorithms for 25 different SFLNs with 20 to 500 nodes were measured and the results are reported in Fig.~\ref{fig:DiabetesTrain}, Fig.~\ref{fig:DiabetesTest} and Fig.~\ref{fig:DiabetesLearntime}, which show that in the simulation of the mid size classification application, ELM can reach a higher testing rate than EELM with same number of nodes. Whereas, the time spent by ELM increases much faster than that spent by EELM with the increasing of the number of nodes.
%% Diabetes Database
\begin{center}
\begin{minipage}[t]{\textwidth}
  \renewcommand{\captionlabeldelim}{.~}
  \renewcommand{\captionfont}{\footnotesize \sffamily}
  \renewcommand{\captionlabelfont}{\footnotesize \sffamily}
  \vspace{0pt}
  \centering
  \includegraphics[scale=0.7]{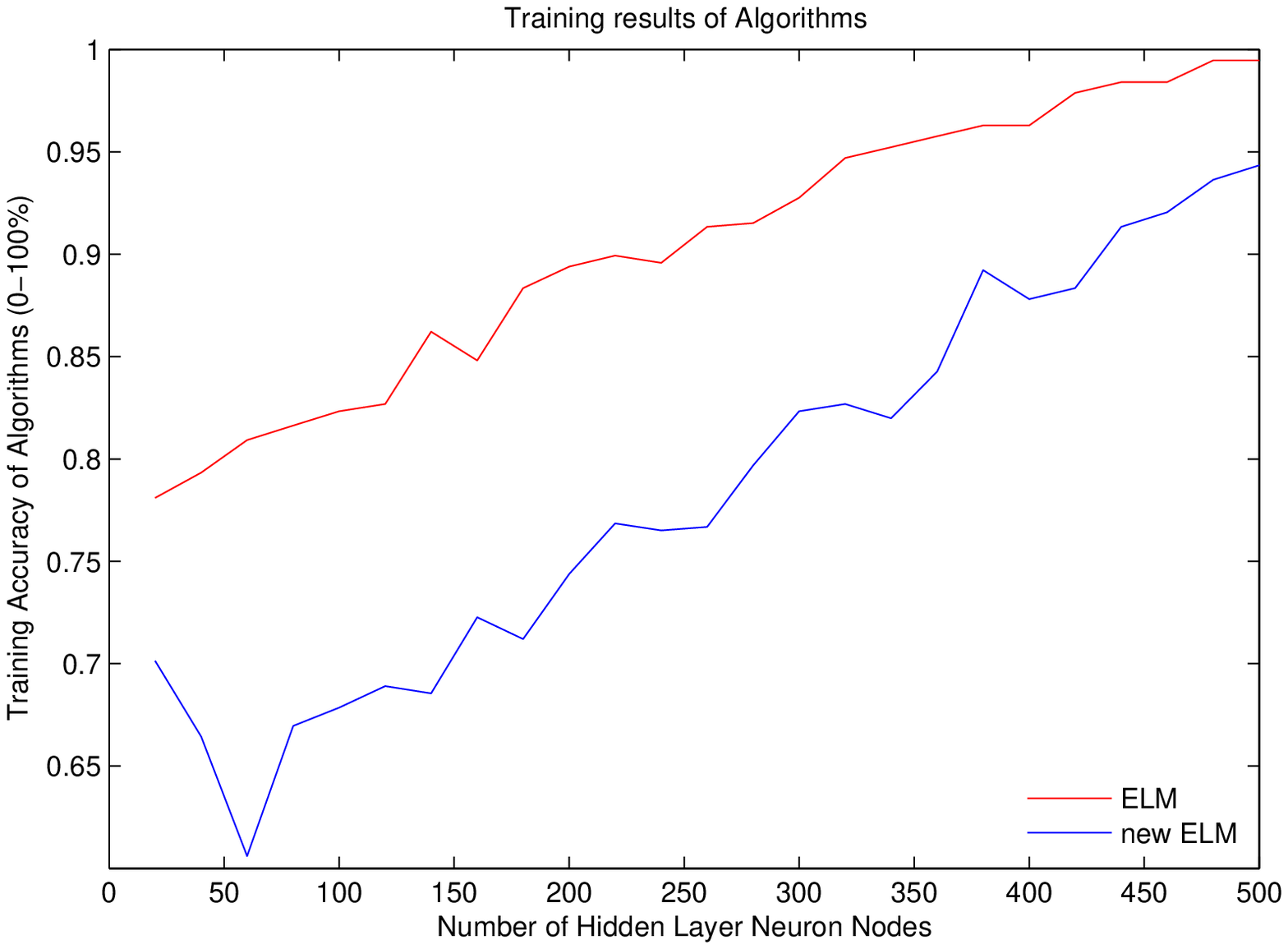}
  \vspace{0ex}
  \renewcommand{\figurename}{Fig.}\vspace{-3ex}
  \figcaption{Training accuracy of two algorithms for Diabetes}\label{fig:DiabetesTrain}
\end{minipage}\hfill
\vspace{5ex}
\begin{minipage}[t]{\textwidth}
   %Requires \usepackage{graphicx}
  \renewcommand{\captionlabeldelim}{.~}
  \renewcommand{\captionfont}{\footnotesize \sffamily}
  \renewcommand{\captionlabelfont}{\footnotesize \sffamily}%\vspace{-1ex}
  \vspace{0pt}
  \centering
  \includegraphics[scale=0.7]{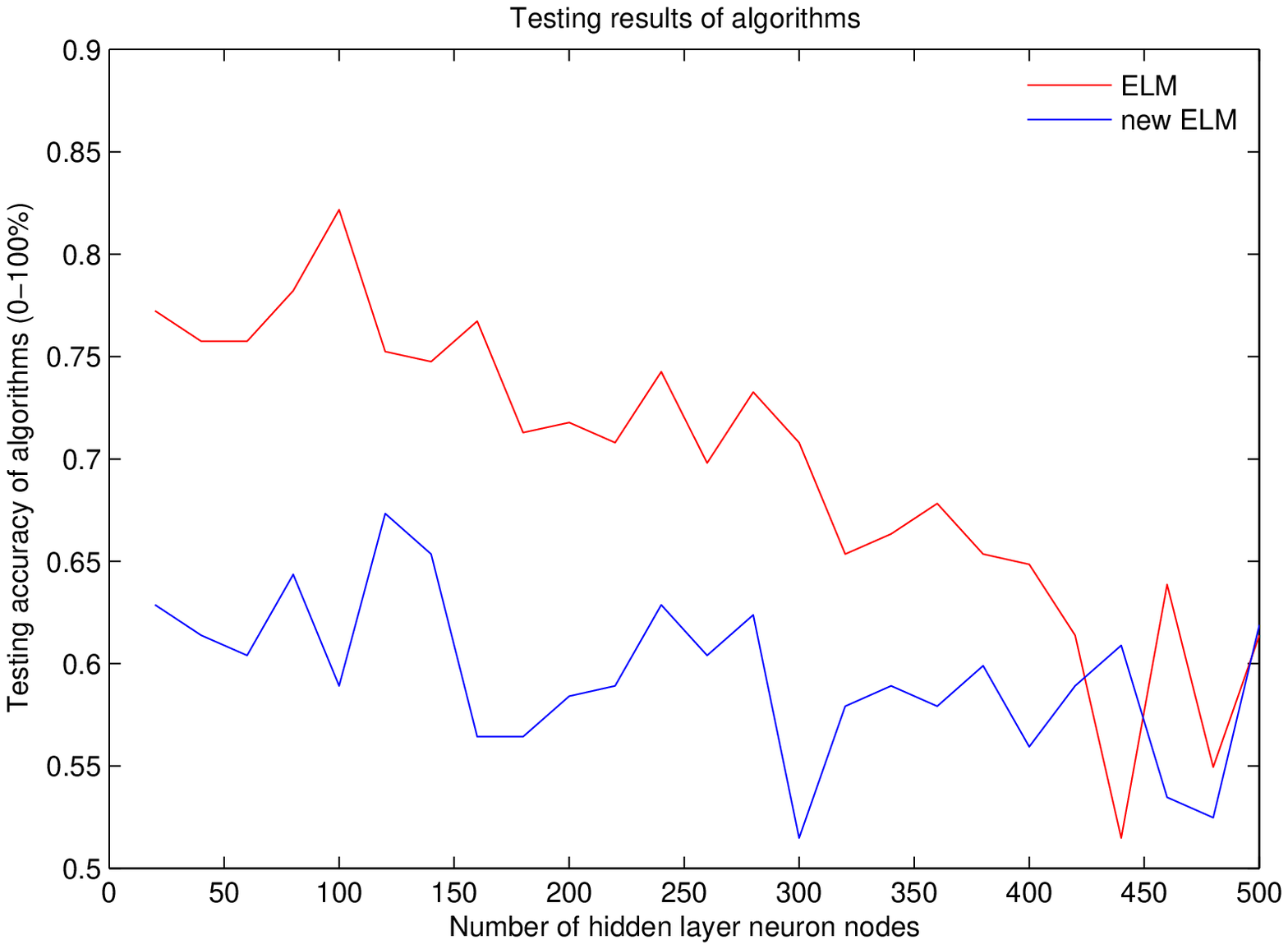}
  \vspace{0ex}
  \renewcommand{\figurename}{Fig.}\vspace{-3ex}
  \figcaption{Testing accuracy of two algorithms for Diabetes}\label{fig:DiabetesTest}
\end{minipage}\hfill
\vspace{5ex}
\begin{minipage}[t]{\textwidth}
   %Requires \usepackage{graphicx}
  \renewcommand{\captionlabeldelim}{.~}
  \renewcommand{\captionfont}{\footnotesize \sffamily}
  \renewcommand{\captionlabelfont}{\footnotesize \sffamily}%\vspace{-1ex}
  \vspace{0pt}
  \centering
  \includegraphics[scale=0.7]{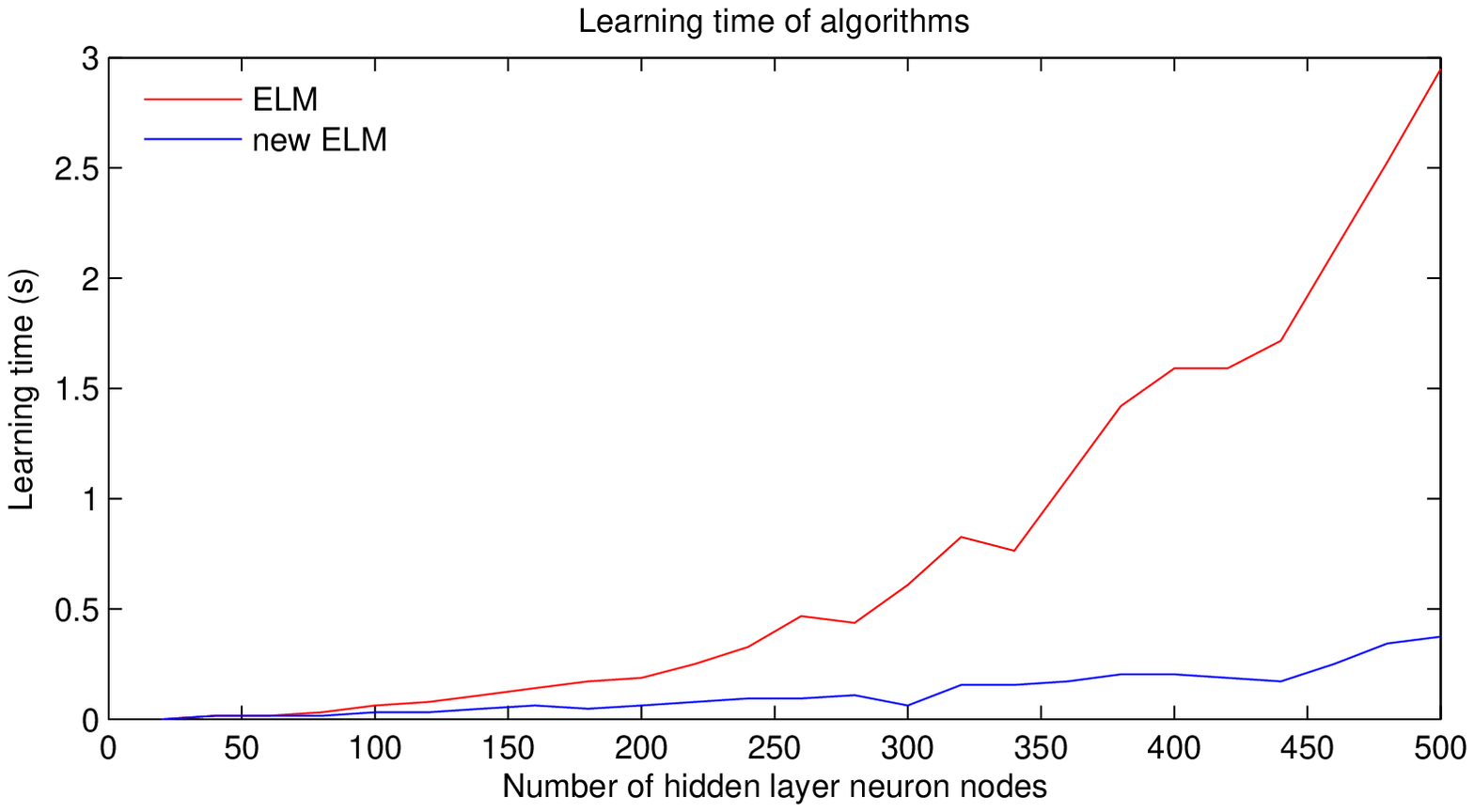}
  \vspace{-0ex}
  \renewcommand{\figurename}{Fig.}\vspace{-3ex}
  \figcaption{Learning time of two algorithms for Diabetes}\label{fig:DiabetesLearntime}
\end{minipage}\hfill
\vspace{1ex}
\end{center}

\section{Discussions and conclusions}
This paper proposed a simple and effective algorithm for single-hidden layer feedforward neural networks (SLFNs) called effective extreme learning machine (EELM) in an attempt to overcome the shortcomings of the extreme learning machine (ELM). There are several interesting features of the proposed EELM algorithm in comparison with the ELM algorithm:
\setcounter{marker}{1}
\begin{list}{(\arabic{marker})}{\usecounter{marker}
\setlength{\labelwidth}{0.55 cm}\setlength{\leftmargin}{0.6 cm}
\setlength{\labelsep}{0.15 cm}\setlength{\rightmargin}{1 cm}
\setlength{\parsep}{0.2ex plus0.1ex minus0.1ex}
\setlength{\itemsep}{0ex plus0.1ex}
\setlength{\topsep}{0.1ex plus0.1ex}}
\item The learning speed of EELM is generally faster than ELM. The main difference between EELM and ELM algorithms lie in the selection of input weights and biases. The ELM algorithm chooses them randomly which consumes little time. Our EELM algorithm selects the input weights and biases properly, which also consumes short time compared with the training time of output weights.
\item The proposed EELM by making proper selection of input weights and biases of the neural networks avoids the risk of yielding singular or not full column rank hidden layer output matrix $\mathbf{H}$. This allows for use of a faster method which can calculate the Moore-Penrose generalized inverse of $\mathbf{H}$ much more rapidly.
\item Another impressive feature the EELM possesses is that it has a longer prediction term with acceptable accuracy than the ELM algorithm. Also, EELM has better robustness property than ELM. In particular, in the regression, the performance of ELM is sometimes poor. But EELM remains steady and has a good performance.
\end{list}

It is worthwhile pointing that in our algorithm in order to sort the samples by affine transformation $X  \mapsto W\cdot X+b$, we adopt the method of decimal numeral system. However, when high-dimensional data is come across and the range of the deviation between samples $|x_{i+1,j}-x_{i j}|$ (the symbols here have the same meanings as in Section~3) is very big, the weights $w_{j}^{2}=w_{j}^{1} 10^{\sum_{p=1}^{j}n_{p}}$ become so large that the computer will treat it as infinity. To resolve the problem, one can use algorithms of large number operation. Whether there exist better methods to sort high-dimensional data effectively and simply by an affine transformation keeps open.

Finally, the proposed EELM algorithm is effective when the activation functions satisfy the assumptions in Theorem~2.3. Gaussian radial basis function belongs to this kind of functions. Nonetheless, the sigmoidal function is not included. This poses a new problem of designing algorithms using other kinds of activation functions, which are as effective and fast as EELM.

\end{document}